\documentclass{article}

\usepackage{microtype}
\usepackage{graphicx}
\usepackage{subfigure}
\usepackage{booktabs} 

\usepackage{hyperref}

\usepackage[accepted]{icml2025}

\usepackage{amsmath}
\usepackage{amssymb}
\usepackage{mathtools}
\usepackage{amsthm}
\usepackage{newtxmath}
\usepackage{enumitem}
\usepackage{epsfig}
\usepackage{graphicx}
\usepackage{url}
\usepackage{tabularx}
\usepackage{boxedminipage}
\usepackage{newtxtext}

\usepackage[capitalize,noabbrev]{cleveref}

\theoremstyle{plain}

\theoremstyle{definition}

\theoremstyle{remark}

\usepackage[textsize=tiny]{todonotes}

\icmltitlerunning{BIBIMBAP: A Benchmark for Instructional BIM-Based Automated Programming}

\begin{document}

\twocolumn[
\icmltitle{BIBIMBAP: A Benchmark for Instructional BIM-Based Automated Programming}



\icmlsetsymbol{equal}{*}

\begin{icmlauthorlist}
\icmlauthor{Clemens Kujat}{ai}
\icmlauthor{Bharathi Kannan Nithyanantham}{vac}
\icmlauthor{Stefan Telgmann}{ai}
\icmlauthor{Tobias Sesterhenn}{tuc}\\
\icmlauthor{Joern Ploennigs}{ai}
\icmlauthor{Stefan Lüdtke}{vac}
\icmlauthor{Christian Bartelt}{tuc}
\end{icmlauthorlist}

\icmlaffiliation{ai}{AI for Sustainable Construction, University of Rostock, Germany}
\icmlaffiliation{vac}{Institute for Visual and Analytic Computing, University of Rostock, Germany}
\icmlaffiliation{tuc}{Institute for Software and Systems Engineering, TU Clausthal, Germany}

\icmlcorrespondingauthor{Clemens Kujat}{clemens.kujat@uni-rostock.de}

\icmlkeywords{Building Information Modeling, Artifical Intelligence}

\vskip 0.3in
]

\printAffiliationsAndNotice{} 

\begin{abstract}
Large Language Models (LLMs) are emerging as natural-language interfaces for Building Information Modeling (BIM), successfully achieving information retrieval, but no benchmark exists to evaluate their ability to edit Industry Foundation Classes (IFC). We introduce BIBIMBAP, a Text-to-BIM benchmark with 100 curated tasks covering Create-Read-Update-Delete (CRUD) operations across spatial, geometric, topological, numeric, and conceptual categories. Each task includes a natural-language prompt, an IFC model, expected structured outputs, and test scripts for automated evaluation. Baseline results show limited performance (best: 50.2\%) and frequent failures in constraint preservation and spatial reasoning, with some correct outputs achieved despite flawed reasoning. Code and data are available at \url{https://nbharathik.github.io/bibimbap/}.
\end{abstract}

\section*{Introduction}

Architectural planning can be seen as an iterative process in which design ideas develop through repeated cycles of interpretation and action rather than through linear refinement \citep{schon1992kinds}. This process ranges from fundamental conceptual revisions to detailed modifications of façades, individual spaces, or technical components. In BIM-based planning (Building Information Modeling) with computer-aided design (CAD) tools, this is reflected in a dynamic workflow in which individual elements or component groups, such as walls, slabs, and openings, are iteratively created, moved, deleted, or duplicated to explore alternatives of design trade-offs. The resulting solutions aim to satisfy functional and architectural requirements and comply with regulatory constraints.

Several approaches have been proposed to automate the design workflow. More recently, LLM-based approaches have been investigated that enable Text-to-BIM generation, e.\,g., \cite{du2024text2bim,deng2025bimgent}. These approaches follow an \textit{automated programming} paradigm in which LLMs generate code from a user prompt to create IFC files. While such one-shot approaches may produce simple models, they do not support the iterative refinement process typical of architectural planning. This has changed with the emergence of tool-augmented LLM frameworks, e.g., using the Model Context Protocol (MCP), which define communication interfaces that allow LLMs to access Application Programming Interfaces (APIs) that tools often provide. LLMs thereby understand which functions are available and how to use them. This unlocks LLMs' access to single CAD tools themselves and, more importantly, enables iterative Text-to-BIM workflows.

\begin{figure*}
    \centering
    \includegraphics[width=0.95\linewidth]{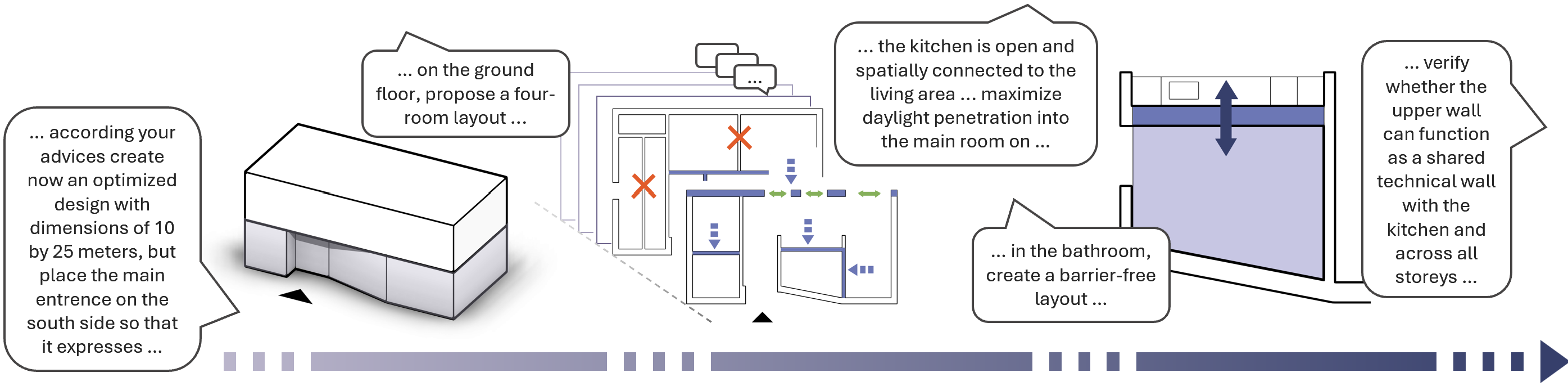}
    \caption{Example of an agentic Text-to-BIM process. Through stepwise interaction with the planner, the LLM provides suggestions, plans changes, and executes the actions in the model, without the planner using the CAD tool itself.}
    \label{fig:design_progress}
\end{figure*}

An example of such a future dialog-oriented assistive workflow is shown in \cref{fig:design_progress}. Within this agent-based Text-to-BIM workflow, planners can iteratively develop design proposals, refine layouts, or elaborate specific design details through natural language interaction. The LLM can not only provide suggestions but also execute incremental modifications to the building model directly. Over successive design revisions, planners can gradually converge toward a coherent design, reflecting the iterative nature of established architectural planning practices.

However, an iterative design process requires new levels of spatial and contextual understanding by the LLM, because it needs to understand the current design context from the targeted object, design intents, and causal implications. It is debated in the literature whether LLMs exhibit these robust geometric, spatial, topological, and numerical reasoning capabilities required for engineering \citep{dong2025ai,ploennigs2025building} and architecture concepts \citep{ching2023architecture}. 

Therefore, it is necessary not only to demonstrate what may be possible but also to identify the current limitations of Text-to-BIM systems and to determine whether these limitations stem from the underlying LLMs or from system-level constraints.

Benchmarks are an important instrument for uncovering these limitations and for establishing baselines against which future research can measure improvements. As discussed, end-to-end test do not serve well the iterative refinement approach as they do not ensure to cover all actions, particular update actions. Therefore, we propose to \textit{benchmark the underlying atomic actions}, as their reliable execution is a prerequisite for any iterative Text-to-BIM workflow. Evaluating whether LLMs can execute individual CRUD actions correctly provides: (i) relevant insights into where and why LLMs fail with simple actions; (ii) allow to recombine those actions into iterative workflows; and (iii) avoid instable actions in such workflows. 

The state-of-the-art analysis in the next section will show that, to the best of the authors' knowledge, no LLM benchmark exists that systematically evaluates all CRUD actions (Create, Read, Update, and Delete) on IFC models.
To address this gap, we present a \textbf{B}enchmark for \textbf{I}nstructional \textbf{BIM}-\textbf{B}ased \textbf{A}utomated \textbf{P}rogramming (\textbf{BIBIMBAP}), which is systematically designed to evaluate Text-to-BIM systems across a curated suite of BIM tasks spanning the full CRUD spectrum. BIBIMBAP emphasizes (i) manipulation tasks that require spatial/geometric and topological reasoning within IFC models, (ii) structured outputs for retrieval tasks that enable direct comparison to ground truth, and (iii) automated tests and evaluation metrics that diagnose where systems fail (e.\,g., existence, location, dimension, and integrity constraints) without relying on human interpretation. By providing an automated benchmark with tasks, reference IFC models, and executable tests, BIBIMBAP aims to make Text-to-BIM research more reproducible and to support objective comparisons of both language models and the BIM tool interfaces they operate through. Due to its atomic nature, the supported BIM tool interfaces, can be low level frameworks, tool APIs as well as MCPs to test all LLM-based automated programming approaches. Therefore, BIBIMBAP's contributions are:
\begin{itemize}[noitemsep, topsep=0mm]
    \item A set of Text-to-BIM tasks spanning the full CRUD operations and multiple reasoning categories.
    \item A prompt grammar (operator/operand/context specifiers) that standardizes task formulation and supports systematic analysis and extensibility.
    \item An automated pipeline with procedural tests and fine-grained evaluation metrics (existence, location, dimensions, correctness, and IFC integrity constraints).
    \item Baseline experiments: A validation of the benchmark across different LLMs. 
\end{itemize}

\section*{Related Work}
This section reviews existing benchmarks and systems for natural-language interaction with BIM, with a focus on how they evaluate retrieval and manipulation tasks. While a growing body of work addresses isolated aspects of Text-to-BIM, like information retrieval, specific generation scenarios, or object recognition, there is currently no widely adopted benchmark that systematically evaluates \emph{all} BIM CRUD operations within a unified framework. Existing benchmarks can be distinguished into: information retrieval benchmarks, datasets related to creation and modification, and end-to-end BIM CRUD approaches. BIBIMBAP is positioned with respect to the remaining gaps across these categories.

\subsection*{Information Retrieval Benchmarks}

\textit{IFC-Bench-v1} provides curated IFC reference models paired with natural-language queries to assess BIM comprehension and reasoning in an agentic NL-to-IFC retrieval setting, primarily covering the \emph{Read} operation \citep{hellin2025nlir}. Tasks are grouped according to whether the requested information is directly, indirectly, or insufficiently available in the model. The released dataset lacks explicit class annotations, which requires manual recategorization, and many tasks rely on open-ended answers evaluated via an LLM-as-a-judge approach, which can be costly and unreliable for free-form responses \citep{wei2024systematic}. In contrast, our benchmark emphasizes structured outputs, explicit evaluation criteria, and a task taxonomy spanning all CRUD operations and multiple reasoning perspectives (e.\,g., geometric and topological), enabling finer-grained and reproducible evaluation.

\textit{BIMCoder} \citep{liuchen2025bimcoder} introduces 1{,}680 retrieval tasks grouped into object-, location-, and relationship-based queries, which largely correspond to basic retrieval scenarios in our taxonomy. Its evaluation is based on exact matching of generated IFCQL queries and execution accuracy. Other retrieval-focused datasets include \textit{BIN-LQ} \citep{wang2022transfer}, which contains 2{,}065 direct questions about element properties or quantities, and \textit{BIM-NLQ} \citep{yin2023ontology}, which supports ontology-aided querying but lacks explicit ground-truth answers, limiting quantitative evaluation. \textit{BuildingQA} \citep{mulayim2025buildingqa} addresses natural-language question answering over building knowledge graphs represented in RDF/TTL format, but it operates on a different data representation and does not target IFC-based manipulation.

\subsection*{Further AI-Driven BIM Benchmarks and Datasets}
Beyond retrieval, several benchmarks address aspects of BIM creation and modification without focusing on Text-to-BIM interaction. \emph{IFCNet} provides single-entity IFC files spanning many IFC classes, enabling controlled evaluation of geometry- and semantics-based methods, but it targets object-level recognition rather than multi-step authoring or manipulation workflows \citep{emunds2021ifcnet}.

In as-built reconstruction, \emph{BIMNet} offers a scan-to-BIM dataset with metrics that explicitly capture geometric and topological reconstruction quality \citep{liu2025bimnet}. Although it is not driven by natural language, its emphasis on explicit constraint checking aligns with the requirements of reliable IFC editing. \emph{ResBIM} contributes a synthetic paired dataset of 2D floor plans and corresponding BIM models for reference-based generation \citep{liang2026resbim}. 

\subsection*{BIM CRUD Approaches}

Text-to-BIM systems translate natural-language intent into executable modeling actions, often via multi-agent decomposition and tool or API calling, enabling the generation of editable, semantically rich building models \citep{du2024text2bim,dong2025ai}. Recent work frames BIM authoring as a computer-use problem: BIMgent employs multimodal LLM agents to plan and execute GUI-level actions in BIM software \citep{deng2025bimgent}. More recently, MCP-based approaches aim to standardize tool access for LLMs to query and edit IFC models directly, as exemplified by \textit{MCP4IFC} \citep{nithyanantham2025mcp4ifc}.

Despite increasing capabilities across Create, Update, and Delete operations, evaluation remains fragmented. Existing studies typically report task-specific success rates or qualitative outcomes, but provide limited standardized, fine-grained diagnostics of geometric correctness, topological validity, and constraint integrity across the full CRUD spectrum, motivating the need for a dedicated benchmark.

\section*{Benchmark}\label{sec:benchmark}
\begin{figure*}[ht]
    \centering
    \includegraphics[width=0.64\linewidth]{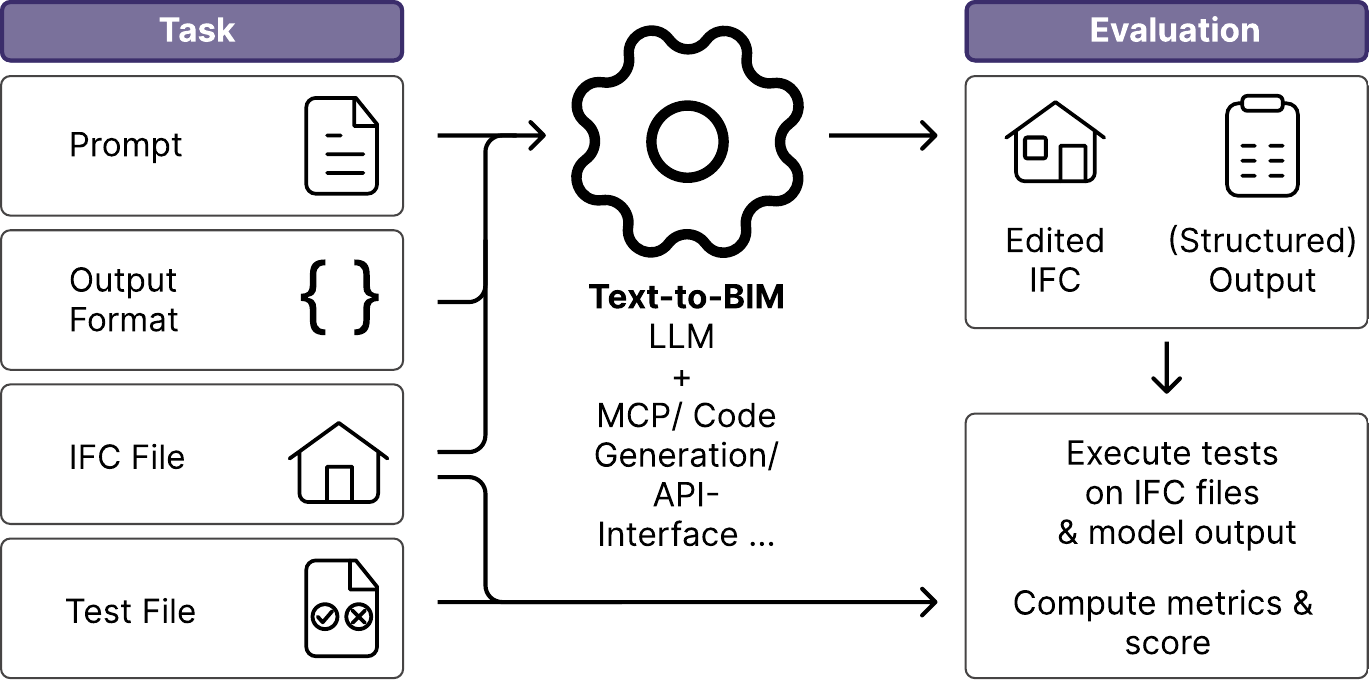}
    \caption{Benchmark architecture: the Text-to-BIM (LLM) takes a prompt and an IFC file (with an output structure for retrieval tasks) and produces an edited IFC file and optional structured output, which are evaluated using the test file and the original IFC model.}
    \vspace{3pt}
    \label{fig:architecture}
\end{figure*}

We introduce \textbf{BIBIMBAP}, a benchmark of 100 tasks for evaluating Text-to-BIM systems on both information retrieval and BIM model manipulation. Tasks are organized along two dimensions: CRUD operation (Create, Read, Update, Delete) to reflect practical BIM workflows, and reasoning requirement, including direct, spatial-geometrical, topological, numeric-calculative, and conceptual tasks. Beyond evaluating LLM capability, BIBIMBAP also evaluates the Text-to-BIM system itself by checking whether the required BIM operations are available and exposed in a way that models can use reliably. To support broad adoption, the benchmark framework provides a unified interface that allows easy integration of different system backends, including direct code-generation pipelines, MCP-based tool interfaces, and custom implementations. The evaluation operates on native IFC, for both input and output, as it is an open standard that most proprietary BIM systems can import and export, making BIBIMBAP applicable to any Text-to-BIM system regardless of implementation details.


\subsection*{Benchmark Architecture}

Each task consists of (i) a natural-language prompt, (ii) a task-specific IFC file containing only the elements needed for the task, (iii) a Python test script that computes evaluation metrics, and (iv) for retrieval tasks, a structured output schema. The IFC files were created specifically for individual tasks and include only the information needed to solve them. At the same time, they are designed so that the task is not trivial; for example, retrieval tasks may include multiple competing candidate objects. \Cref{fig:architecture} illustrates the execution flow. Tasks are executed independently by design, as they represent atomic steps within an iterative design process, such as creating a single wall, inserting a door, moving one element, or retrieving one quantity. This is a deliberate methodological choice: the benchmark does not aim to evaluate the solving of complete multi-step design objectives, such as generating an entire house from a single prompt, but rather the model's ability to reliably perform individual, well-defined atomic actions that are prerequisites to reliably fulfill multi-step objectives.

For automated evaluation, the system prompt discourages the LLM from asking clarification questions. The benchmark framework allows users to configure both the evaluated LLM and the number of samples per task. All model outputs are logged, and modified IFC files are stored for Create, Update, and Delete tasks. Evaluation is deterministic and depends on the task type: retrieval tasks expect structured outputs from the LLM with explicit schemas for direct comparison with ground truth, avoiding LLM-as-a-judge evaluation \cite{hellin2025nlir}. Create, Update, and Delete tasks use task-specific test scripts written in Python, using IfcOpenShell, which analyze the original and modified IFC files to compute the metrics. Each test script was validated by a second person to assure correctness.

\subsection*{Task Categorization}\label{sec:task_categorization}

As Text-to-BIM systems do not fail uniformly across different tasks, we use distinct categories to systematically separate different error modes, enabling not only to diagnose that tasks fail, but rather understand failure origins. The benchmark is therefore structured into five main categories: direct tasks, spatial-geometrical reasoning, topological reasoning, numeric-calculative reasoning, and conceptual understanding. Each category contains a total of 20 tasks. The first three categories are subdivided according to the four CRUD operations, with five tasks assigned to each one, while the numeric-calculative and conceptual understanding categories consist of 20 retrieval tasks each. By structuring the benchmark along these categories, we make explicit which aspects of BIM understanding current LLM-based systems struggle with, rather than reporting a single aggregated performance score, thereby enabling to assess a models specific strengths and limitations.

\subsubsection*{Task Complexity}
The general difficulty of individual task categories results from the information provided in the prompt. Direct tasks contain explicit information about the targeted elements, so that the LLM can identify them concretely by their ID. Geometric-spatial, topological, numeric-calculative and conceptual tasks contain a prompt component with implicit information about the targeted elements. Based on a contextualized description (e.g. "element that is enclosed by"), the LLM has to derive the targeted elements through model properties and spatial relationships. Across all CRUD operations, domain-specific contexts, properties and expressions are applied to address realistic and practice-oriented tasks. This is particularly evident in numeric-calculative and conceptual tasks: Numeric-calculative tasks require mathematical reasoning and domain-specific knowledge for their solution, while conceptual tasks require the capacity to assess abstract architectural relationships.

\subsubsection*{Direct Tasks}
These tasks test LLM capabilities for basic BIM operations and examine whether fundamental planner intentions can be understood and correctly executed without requiring complex reasoning. Each prompt includes all required details (e.\,g., element IDs, coordinates, and dimensions), so the task can be solved directly, as all information for understanding the scene is explicitly provided in the prompt. Examples include moving objects within a local coordinate system, deleting objects specified by explicit element IDs, and adding objects with numerically defined dimensions, placement, and orientation (e.\,g., ``Add a wall of length 15 m starting at (0, −10)``).

\subsubsection*{Spatial-Geometrical Tasks}
Spaces are configurations of surfaces, volumes, and openings whose effects emerge through placement, orientation, scale, and angular relations, and in building design these configurations determine how spaces, functionality, and spatial perception are produced. Spatial-geometrical tasks therefore test whether an LLM can reason about such spatial and geometric relationships. The model must identify the correct elements based on shape cues (e.\,g., ``the room that is not a perfect square``) by computing properties such as lengths, perimeters, or bounding-box dimensions. It should also follow spatial instructions (e.\,g., ``the door between doors X and Y``) by understanding relations such as between, next to, or opposite. These tasks assess whether an LLM actually reasons about the spatial alignment and require the model to construct an internal spatial representation and derive properties such as relative position, betweenness, and shape characteristics.


%

\subsubsection*{Topological Tasks}
Topological reasoning tasks test whether an LLM can understand functional and technical relationships such as adjacency, enclosure, openings, connectivity, and contact. In building design, these topological relations generate functionality and spatial organization, for example, whether a wall fully separates a kitchen from a living room or defines a shared space. While some relations are explicitly stored in IFC (e.\,g., a column assigned to the space it belongs to), many important relations are not represented directly and must be inferred through higher-level topological reasoning. These tasks, therefore, require the model to reason over collections of elements, such as identifying how multiple walls jointly form an enclosure or determining which walls separate interior from exterior. Examples include creating a connection between an interior space and the outside by adding a window.


\subsubsection*{Numeric-Calculative Tasks}
In building design and planning, a robust understanding and mathematically correct handling of quantities, areas, volumes, and perimeters is essential, as building regulations frequently specify numerical constraints, and such quantities form the basis for cost estimation, and rentable area calculations used in economic evaluations. These tasks require the LLM to retrieve numeric properties from IFC elements and perform multi-step calculations to produce a final result. They often involve selecting the correct elements (by ID or filter), choosing the appropriate measurement (net vs. gross), aggregating across multiple elements, and applying deductions such as subtracting openings from wall surface areas. As such, they test multi-step reasoning rather than direct value lookup. For example, ``Provide the total net wall area of walls A, B, and C`` requires extracting each wall's surface area after accounting for openings and then summing the values. 

\subsubsection*{Conceptual Tasks}
Architectural design builds on an understanding of how architectural concepts emerge from the interplay of spatial, geometric, and topological properties. For example, functionality is often produced through the organization and hierarchization of spaces, while spatial perception can arise from geometric form, placement, and light. These tasks, therefore, require an understanding of formal and functional architectural concepts that emerge from spatial configuration rather than from explicitly stored properties. Guided by principles of architectural perception and design \citep{ching2023architecture}, we distinguish four categories: spatial--geometric relationships, organizational structures, openings and light, and ordering principles. Solving these tasks often requires combining multiple types of reasoning described earlier. For instance, identifying a corridor requires understanding orientation, proportion, and connectivity patterns, even if the space is not explicitly labeled.

\subsection*{Prompt Construction}
\begin{figure*}[ht]
    \centering
    \includegraphics[width=0.9\linewidth]{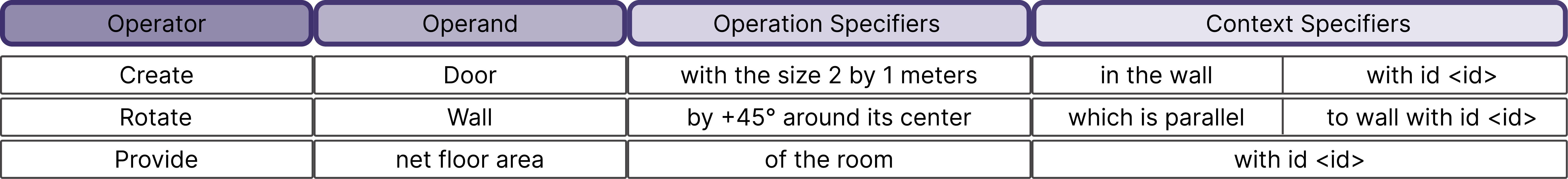}
    \caption{Visualization of the prompt structure using the grammar with three examples.}
    \label{fig:prompt-taxonomy}
\end{figure*}

To ensure a high CRUD task coverage and systematic testing of LLM reasoning capabilities, all prompts follow a structured grammar (\cref{fig:prompt-taxonomy}) that allows to create variants of prompts. Each prompt includes an \textit{operator} that denotes the intended operation (e.\,g., create, rotate, provide); an \textit{operand} that denotes the target building elements or the desired information subject to the task; \textit{operation specifiers} that constrain the request (e.\,g., required properties, relations, or filters), and \textit{context specifiers} that specify the relevant context elements either implicitly or explicitly. This allows similar tasks to address different reasoning categories by replacing or omitting individual parts of the prompt. For example, a task could be direct with the explicit context specifier "with <id>", while by using the context specifier "which is parallel to wall with <id>", the task would be a spatial-geometrical one.

\subsection*{Evaluation Metrics} 

To quantify performance, we define evaluation metrics. Because tasks differ, it is impractical to define a fixed set of metrics for all of them. Therefore, we define a set of metrics for each CRUD category. The set of metrics and their assignment to task categories is described in \cref{tab:evaluation-metrics}. For each task $t$, we calculate $\operatorname{score}(t)$, and the overall benchmark score is the average of $\operatorname{score}(t)$ over all tasks.

\[\operatorname{score}(t)=\frac{\text{no. of metrics satisfied for }t}{\text{no. of metrics evaluated for }t}.\]

\vspace{3pt}

\begin{table}[h]
\centering
\caption{Evaluation metrics and applicable task categories.}
\label{tab:evaluation-metrics}
\vspace{3pt}
\small
\setlength{\tabcolsep}{6pt}
\renewcommand{\arraystretch}{1.15}

\begin{tabularx}{\columnwidth}{@{}>{\raggedright\arraybackslash}p{0.30\columnwidth} X >{\centering\arraybackslash}p{0.23\columnwidth}@{}}
\toprule
\textbf{Metric} & \textbf{Description} & \textbf{Categories} \\
\midrule
Object exists &
Required object exists in the edited IFC file &
Create \\

Object does not exist &
Referenced object does not exist in the edited IFC file &
Delete \\

Correct location &
Object of interest is in the correct location &
Create, Update \\

Correct dimension &
Object of interest has the correct size &
Create, Update \\

Integrity assured &
No constraints are violated &
Create, Update, Delete \\

Correct answer &
Model provided the correct answer &
Retrieval \\
\bottomrule
\end{tabularx}

\vspace{3pt}
\end{table}

These metrics provide more insights into where an LLM fails, rather than simply determining whether a task was completed correctly. For example, the task “Create a door with the size of 2 by 1 meters in the wall” is evaluated using the metrics \emph{Object exists}, \emph{Correct location}, \emph{Correct dimension}, and \emph{Integrity assured} (a corresponding opening must exist that voids the wall and is filled by the door). The integrity metric is task-specific. For example, inserting a wall to enclose a room requires topological relationships to the other walls. Deleting an element requires that IFC relationships (e.\,g., \textit{IfcRelContainedInSpatialStructure}) are also removed. Furthermore, the integrity metric verifies that no clashes are introduced, the spatial hierarchy is preserved, and non-target elements remain unchanged. Retrieval tasks use \emph{Correct answer}, comparing the structured output against ground truth.  

\section*{Experiments}

We evaluate Claude Opus 4.6, GPT-5.4, Gemini 3 Flash, GPT-5.2, GPT-5 Mini, and GPT-4.1 in a minimal tool-based setup implemented with the LangChain framework. Each model is given access to a single tool, \texttt{execute\_ifc\_code}, which runs Python code with IfcOpenShell in a sandboxed environment. The models may call this tool multiple times to solve a task. To prevent infinite loops, we limit the number of tool calls at 25 per task. No task-specific examples or hints are provided, which ensures a fair baseline setting. All models use the same system prompt, agent framework, and tool access. To demonstrate that the top-performing models produce consistent results across runs, we run the three newest models (Claude Opus 4.6, GPT-5.4, and Gemini 3 Flash) three times each and report mean scores with standard deviations. The remaining three models (GPT-5.2, GPT-5 Mini, and GPT-4.1) are evaluated with a single run. 

\subsection*{Code-Execution Pipeline}

The benchmark evaluates LLMs through a code-execution pipeline rather than direct IFC text editing. When the LLM receives a task prompt, it generates Python code that uses the IfcOpenShell library. This code is then passed to the \texttt{execute\_ifc\_code} tool, which runs it in a sandboxed subprocess with the IFC file already loaded as the variable \texttt{ifc}. For editing tasks, the LLM must call \texttt{commit()} to save the changes. Geometry is modified through the IfcOpenShell API, for example by creating representations with \texttt{ifcopenshell.api.run()}. The LLM may make several tool calls in sequence, first to inspect the model and then to apply the required changes. For example, consider the prompt: \emph{``Delete all doors on the first floor.''} The LLM may first generate code to inspect the IFC model and find which storey corresponds to the first floor and which doors belong to it. After receiving the result, it then generates a second code snippet to delete those doors and calls \texttt{commit()} to save the changes. All the generated code is executed through the \texttt{execute\_ifc\_code} tool. 

Our goal is to establish strong baseline results that can serve as a reference point for future improvements. Failures on particular task types highlight open challenges (e.\,g., geometric reasoning, IFC relationship handling, and BIM-aware representations), while strong performance indicates tasks that current LLMs already support reliably.

\section*{Results}
\Cref{tab:model_category_overall} summarizes the overall model performance with a breakdown by category.
Overall, the newer models achieve higher scores; however, the performance still remains low, which shows that Text-to-BIM manipulation is still challenging for an LLM. Claude Opus 4.6 achieves the best overall average score (50.2\%), followed by GPT-5.4 (31.3\%), Gemini 3 Flash (30.8\%), GPT-5.2 (30.5\%), GPT-5 Mini (25.0\%), and GPT-4.1 (23.9\%). The top three models show consistent results across repeated runs. Even the best model solved only about half the tasks on average, which highlights the difficulty of reliably using LLMs for CRUD and fulfilling all evaluation criteria.


\begin{table*}[h]
\centering
\caption{Model performance by category and overall average score (\%). The top three models are averaged over three runs ($\pm$ std.\,dev.). Models marked with * are single-run results.}
\label{tab:model_category_overall}
\vspace{3pt}
\small
\setlength{\tabcolsep}{6pt}
\renewcommand{\arraystretch}{1.05}
\begin{tabularx}{\textwidth}{@{}X *{5}{@{\hspace{18pt}}>{\centering\arraybackslash}r} @{\hspace{16pt}}>{\centering\arraybackslash}r@{}}
\toprule
\textbf{Model} &
\textbf{\shortstack{Direct (20)}} &
\textbf{\shortstack{Spatial (20)}} &
\textbf{\shortstack{Topological (20)}} &
\textbf{\shortstack{Numeric (20)}} &
\textbf{\shortstack{Conceptual (20)}} &
\textbf{\shortstack{Overall Score}} \\
\midrule
Claude Opus 4.6 & 65.83 $\pm$ 0.16 & 63.89 $\pm$ 0.22 & 53.06 $\pm$ 0.19 & 38.33 $\pm$ 0.24 & 30.0 $\pm$ 0.21 & \textbf{50.22 $\pm$ 0.22 } \\
GPT-5.4  & 62.15 $\pm$ 0.40 & 42.08 $\pm$ 0.24 & 20.56 $\pm$ 0.10 &  26.67 $\pm$ 0.22 & 5.00 $\pm$ 0.00 & 31.29 $\pm$ 0.04 \\
Gemini 3 Flash & 65.42 $\pm$ 10.4 & 40.97 $\pm$ 1.73 & 21.11 $\pm$ 3.37 & 20.00 $\pm$ 5.00 & 6.67 $\pm$ 2.33 & 30.83 $\pm$ 1.91 \\
\midrule
GPT-5.2*             & 52.50 & 34.58 & 35.42 & 5.0 & 25.0 & 30.50 \\
GPT-5 Mini*          & 50.94 & 25.83 & 25.83 & 0.0 & 25.0 & 25.00 \\
GPT-4.1*             & 38.33 & 30.00 & 31.25 & 0.0 & 20.0 & 23.92 \\
\bottomrule
\end{tabularx}
\end{table*}

Performance differences across categories show clear trends. Direct tasks are relatively easy because they often provide explicit identifiers, coordinates, or dimensions. Spatial-geometrical and topological tasks are harder because they require selecting the correct target elements from context and reasoning about relationships or placements. Numeric-calculative tasks are the most difficult for all models, with consistently low scores. These tasks typically require aggregating information across multiple elements and computing quantities from geometry (e.\,g., areas, volumes, and totals). Models often produce partially correct code but miss elements, apply an incorrect quantity definition, use wrong unit scale or return an incorrect value. The consistently low scores on numeric-calculative tasks suggest that BIM reasoning remains a major limitation for the used LLMs.


Tool usage patterns differ across models. All models rely heavily on the tool interface to solve tasks. Claude Opus 4.6 makes the most tool calls, averaging around 4.9 per task, followed by Gemini 3 Flash with about 3.5 calls, GPT-4.1 with 2.7 calls, GPT-5 Mini with 2.5 calls, GPT-5.2 with 2.2 calls, and GPT-5.4 with only 1.67 calls. Notably, Claude Opus 4.6 achieves the highest overall score while also using the most tool calls, suggesting that its iterative strategy contributes to its stronger performance. 

Table \ref{tab:crud_model_share} shows the performance of the LLMs by CRUD operation. Create is the easiest operation in the current benchmark, since most creation tasks focus on simple building elements with explicit parameters. Claude Opus 4.6 reaches 81.1\% on Create tasks. Retrieve is the most difficult because it requires both reliably identifying the correct target elements and computing or formatting the answer. Update tasks require precise geometric modifications, while Delete tasks can appear easy, but many failures arise from integrity requirements: deleting an element often requires removing related relationships or associated elements (e.\,g., when a door is deleted, its associated opening also needs to be deleted). As a result, a model may remove the main element but still fail the task if the IFC remains inconsistent.

\begin{table}[h]
\centering
\caption{CRUD success rate by model (\%).}
\label{tab:crud_model_share}
\vspace{2pt}
\small
\setlength{\tabcolsep}{3.5pt}
\renewcommand{\arraystretch}{1.0}
\begin{tabularx}{\columnwidth}{@{}Xcccc@{}}
\toprule
\textbf{Model} & \textbf{Create} & \textbf{Read} & \textbf{Update} & \textbf{Delete} \\
\midrule
Claude Opus 4.6 & 81.1$\pm$0.2 & 47.3$\pm$0.3 & 61.5$\pm$0.2 & 18.9$\pm$0.1 \\
GPT-5.4         & 56.0$\pm$0.3 & 22.4$\pm$0.0 & 50.4$\pm$0.3 & 20.0$\pm$0.2 \\
Gemini 3 Flash  & 45.6$\pm$8.4 & 20.6$\pm$1.1 & 46.7$\pm$2.2 & 37.8$\pm$6.9 \\
\midrule
GPT 5.2*         & 58.9 & 18.2 & 51.1 & 26.7 \\
GPT-5 Mini*      & 31.2 & 10.9 & 55.5 & 43.3 \\
GPT 4.1*      & 29.4 &  9.0 & 60.0 & 36.6 \\
\bottomrule
\end{tabularx}
\vspace{2pt}
\end{table}

\begin{figure}[h]
  \centering
  \includegraphics[width=\linewidth]{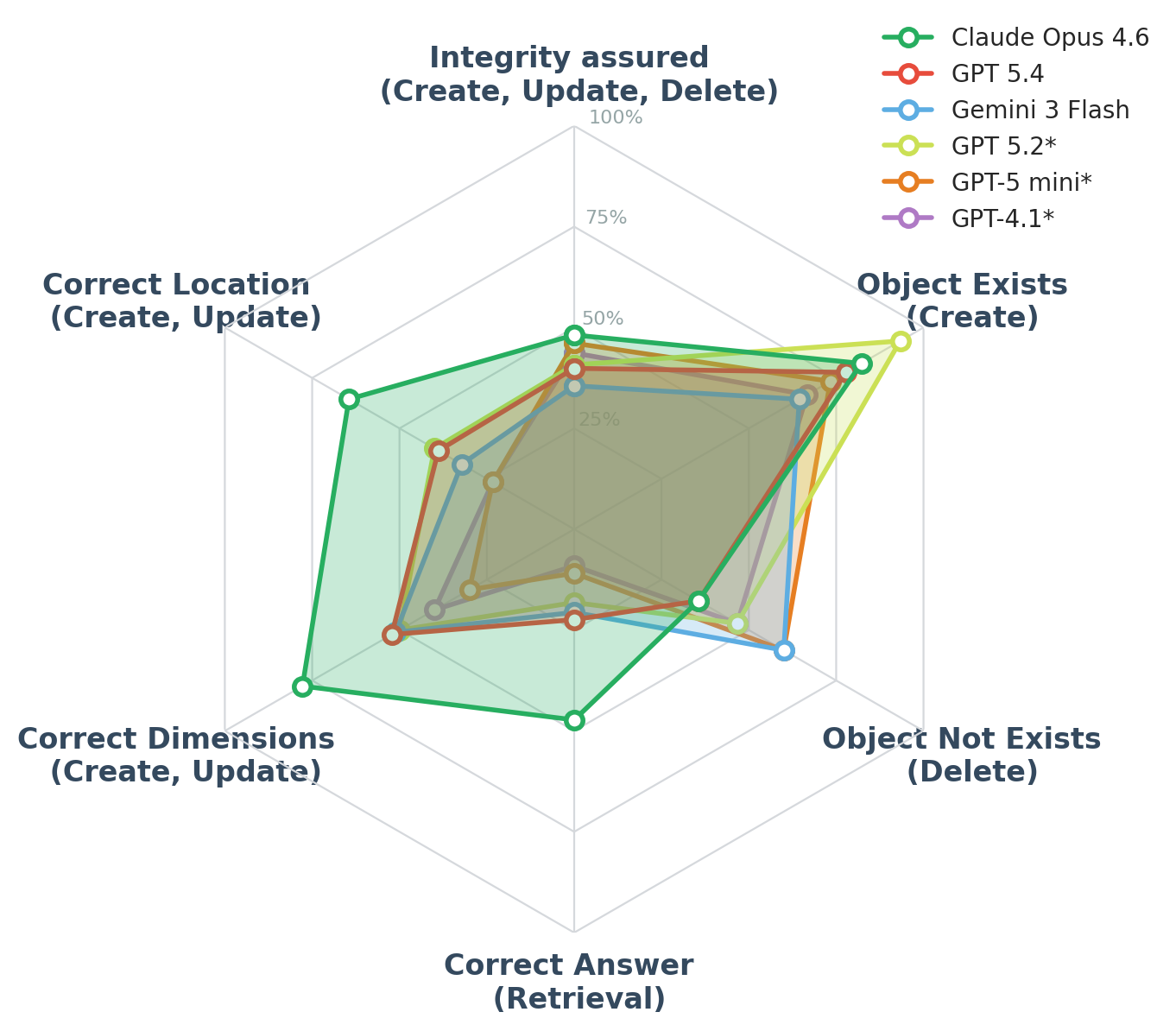}
  \caption{Each model's performance across the six evaluation criteria.}
  \vspace{3pt}
  \label{fig:constraints}
\end{figure}

Figure \ref{fig:constraints} plots the accuracy across the evaluation metrics and highlights a consistent limitation: \emph{integrity\_constraint} is the most challenging criterion for all models. In contrast, the \emph{object\_exists} and \emph{object\_not\_exists} constraints achieve higher scores, indicating that models can often create or delete an entity but still struggle to maintain correct IFC relationships and constraints. Similarly, \emph{correct\_location} and \emph{correct\_dimensions} remain low for most models, which is critical for practical BIM authoring because a created element is only useful if it is placed correctly and has the correct dimensions. For retrieval tasks, \emph{correct\_answer} remains low, suggesting that even when the model can access the relevant elements, it often fails to compute or properly format the final answer.

Overall, these results highlight both the promise and the limitations of current LLMs. Current models can often perform basic Create and Delete actions, and larger, stronger models benefit from iterative tool use. However, robust BIM manipulation requires more than producing valid code: it also requires correct element selection, correct computations, and consistent IFC integrity constraints. The results show, that no model manages to address all those aspects.

\section*{Examples}

To illustrate task complexity, we give one example each for success and failure cases. For each, we show the IFC models alongside the corresponding prompts, as well as the resulting IFC models.

\begin{figure}[h]
  \centering
  \includegraphics[width=\linewidth]{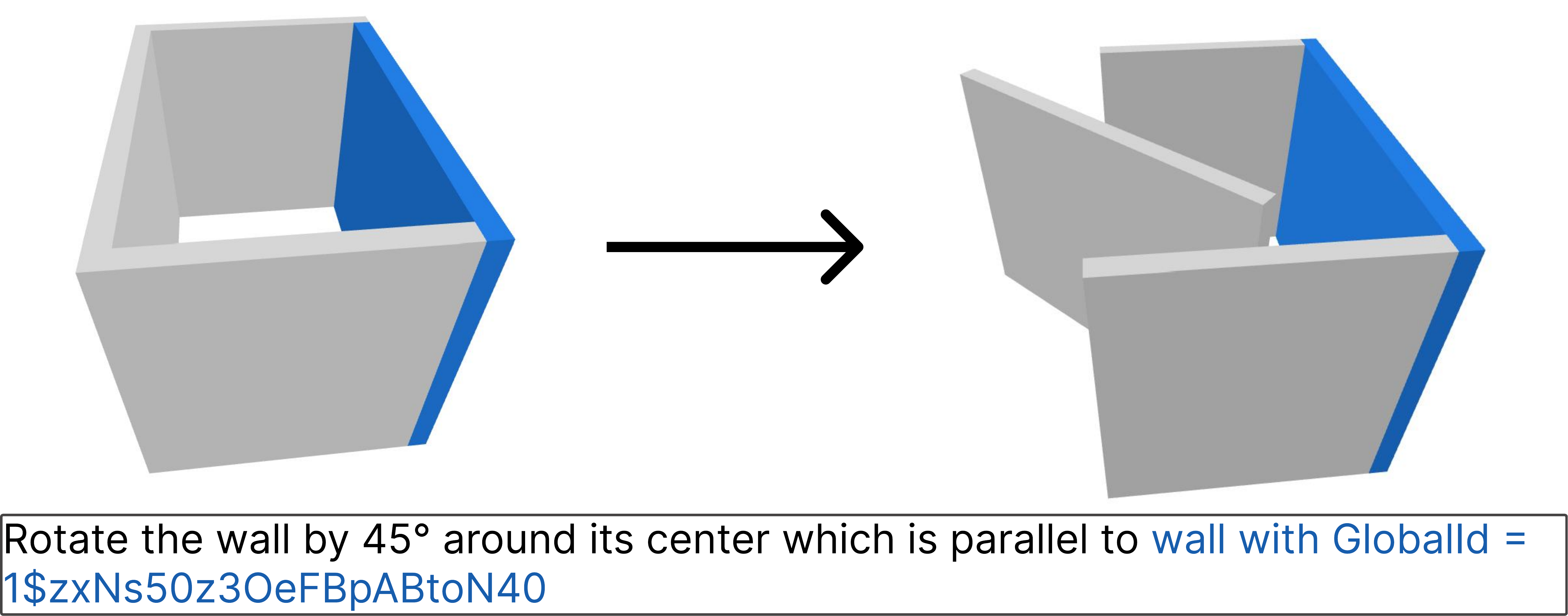}
  \caption{Successfully solved spatial-geometrical task of Claude Opus 4.6}
  \label{fig:success-rotation}
\end{figure}

\Cref{fig:success-rotation} shows a spatial-geometrical task that was successfully solved by Claude Sonnet 4.5. The main challenges are identifying the correct wall to rotate and rotating it about its center while ensuring the correct placement after rotation.
The LLM first extracted the direction of the reference wall. It then extracted the directions of the other walls to find the one with the same direction, computed its rotation center, and finally derived the new direction and location. This process required six tool calls in total.

\begin{figure}[h]
  \centering
  \includegraphics[width=\linewidth]{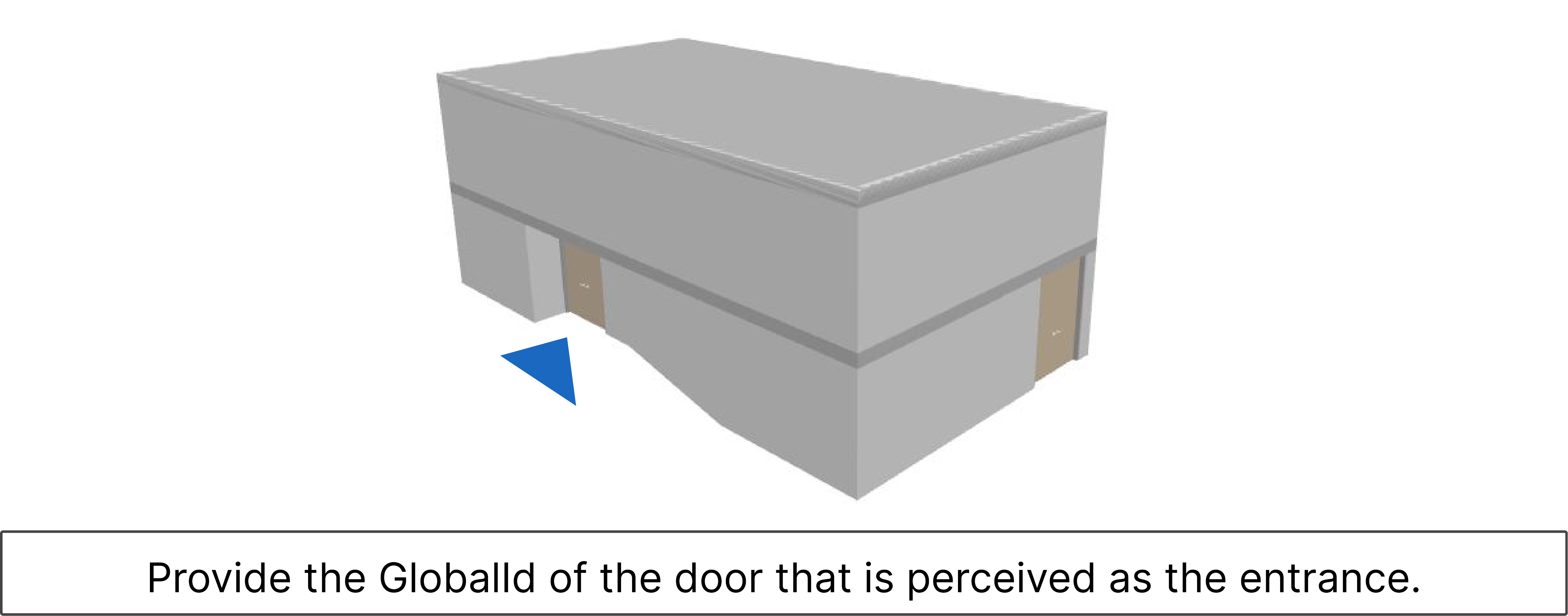}
  \caption{Example of a conceptual task. GPT-5.4 solved it by choosing the right door randomly.}
  \vspace{3pt}
  \label{fig:success-right-guess}
\end{figure}

\Cref{fig:success-right-guess} shows an example of a conceptual task. The task is to identify the door that is perceived as the entrance only by reasoning about the architectural concept.
GPT-5.4 tried to find keywords in door objects that indicate their use as the entrance. Because no such keyword was present, the model selected the first door as the answer, which happened to be correct by chance, even though the model was not able to solve the task in a principled way.
In other tasks in this category, the LLMs exhibited similar issues and produced errors after making too many tool calls.

Another common failure was incorrect unit handling: models sometimes used wrong units or returned incorrect results despite explicit unit specifications.

\section*{Limitations}

The current version of the benchmark consists of a limited number of 100 prompts. Future versions aim to expand the task pool to improve statistical robustness and broader coverage.
Although the benchmark covers all four CRUD operations, retrieval tasks represent the largest share. This is partly by design: the numeric-calculative and conceptual understanding categories consist exclusively of retrieval tasks, and providing quantifiable, directly evaluable answers for these is itself a contribution of this work. Future versions of the benchmark will increase the proportion of Create, Update, and Delete tasks to achieve a more balanced distribution across operation types.

The IFC files used in the evaluation are generally applicable to other tasks, but they primarily serve as task-specific test environments that include only the elements required for the respective task. We therefore selected small models to keep the benchmark focused on individual tasks and to ensure a clear evaluation. However, this also means the benchmark does not fully reflect realistic BIM authoring settings, which are typically larger and noisier and exhibit higher structural complexity (e.\,g., more storeys and a wider variety of building elements).

Our tasks are instructions for atomic operations. In real use, users often interact with an AI system over multiple turns to refine requests and refer to earlier actions (e.\,g., “Undo the last change” or “Now make that window larger”). Those interaction sequences are not part of this work, as models already fail on simple steps.

The evaluation metrics are detailed, but they do not cover all aspects of the BIM authoring standards. The benchmark focuses on hard constraints (geometry, existence, and relationships), but does not evaluate, e.g.\,  whether element naming conventions are followed. We also do not test whether non-geometric properties are preserved in Update tasks; for example, deleting an element and recreating it can count as a correct update even if properties are lost.

\section*{Conclusion}

This paper introduces BIBIMBAP, a benchmark for evaluating Text-to-BIM systems on IFC-based tasks spanning the full CRUD spectrum (Create, Read, Update, Delete) across multiple reasoning categories (direct, spatial-geometrical, topological, numeric-calculative, and conceptual). We report baseline results for several state-of-the-art LLMs in a tool-based setup. Our results indicate that current models can often complete simple actions when parameters are explicit, but they frequently fail on tasks that require geometric or topological reasoning, multi-step computation, or preserving IFC relationships and constraints. These findings highlight that reliable BIM manipulation remains challenging and that progress should be measured with reproducible tests. Overall, BIBIMBAP provides an automated evaluation framework for Text-to-BIM approaches, identifying failure modes and tracking progress over time. We hope it supports both researchers and practitioners in developing more reliable BIM assistants and motivates future work on BIM-aware LLMs and domain-specific training.

\nocite{langley00}

\bibliography{references}
\bibliographystyle{icml2025}




\end{document}